\documentclass[11pt,a4paper]{article}
\usepackage[hyperref]{naaclhlt2019}
\usepackage{times}
\usepackage{latexsym}
\usepackage{graphicx}

\usepackage{url}
\usepackage{algorithm}
\usepackage{algorithmic}

\usepackage{soul}
\usepackage{amssymb}
\usepackage{contour}

\aclfinalcopy 

\title{A Consolidated System for Robust Multi-Document Entity Risk Extraction and Taxonomy Augmentation}

\author{Berk Ekmekci, Eleanor Hagerman and Blake Howald \\
  Thomson Reuters Special Services, LLC\\
  1410 Spring Hill Road, Suite 301 \\
 Mclean, VA 22102-3058\\
  {\tt firstname.lastname@trssllc.com}  }
\date{}

\begin{document}
\maketitle
\begin{abstract}
We introduce a hybrid human-automated system that provides scalable entity-risk relation extractions across large data sets. Given an expert-defined keyword taxonomy, entities, and data sources, the system returns text extractions based on bidirectional token distances between entities and keywords and expands taxonomy coverage with word vector encodings. Our system represents a more simplified architecture compared to alerting-focused systems - motivated by high coverage use cases in the risk mining space such as due diligence activities and intelligence gathering. We provide an overview of the system and expert evaluations for a range of token distances. We demonstrate that single and multi-sentence distance groups significantly outperform baseline extractions with shorter, single sentences being preferred by analysts. As the taxonomy expands, the amount of relevant information increases and multi-sentence extractions become more preferred, but this is tempered against entity-risk relations become more indirect. We discuss the implications of these observations on users, management of ambiguity and taxonomy expansion, and future system modifications.
\end{abstract}

\section{Introduction}

Identifying or predicting entity-risk relationships in textual data is a common natural language processing (``NLP") task known as \textit{risk mining} \cite{LeidnerSchilder2010}. For example, (1) relates the entity \textbf{CNN} to the risk \underline{\smash{pipe bomb}}, which is part of a broader \textit{terrorism} risk category.\\ 

\footnotesize
\noindent (1)\textit{Later Wednesday, \textbf{CNN} received a \underline{\smash{pipe bomb}} at its Time Warner Center headquarters in Manhattan.}\\

\normalsize
\noindent Monitoring systems seek to classify such text extracts as indicating a risk or not for a range of entities and risk categories. Any collection of additional information (e.g., time, location) is often secondary to risk classification. Thus for use cases seeking more coverage - e.g, intelligence gathering, legal and financial due diligence, and operational risk management - our experience is that traditional risk mining architectures are not entirely fit for purpose. Consequently, we present a hybrid-automated system that leverages predefined entities, data sources and risk taxonomies to return extractions based on bidirectional entity-risk keyword surface distances. \\ 

\footnotesize \noindent (2) \textit{On Monday, a \underline{\smash{pipe bomb}} was found in a mailbox of billionaire business magnet and political activist George Soros. Later Wednesday, \textbf{CNN} received a \underline{\smash{pipe bomb}} at its Time Warner Center headquarters in Manhattan sent to ex-CIA director John Brennan and a \underline{\smash{\underline{suspicious package}}} sent to Rep. Maxine Waters ....}\\

\normalsize
By starting with expert defined information, our system can return both (1) and (2) by varying the distance threshold without the need for a risk classification engine, morphosyntactic parsing or named entity recognition. Further, taxonomy coverage is expanded with word vector encodings trained on the same sources of data, e.g., \underline{\smash{\underline{suspicious package}}} in (2). We believe this approach best combines the efficiencies that tuned Machine Learning (``ML") systems can offer with the rich depth of experience and insight that analysts bring to the process.

In this paper, we review risk mining research (Section 2) to contextualize the presentation of our system (Section 3) and expert evaluations testing a range of different distance thresholds with seed and expanded taxonomies across \textit{cybersecurity}, \textit{terrorism} and \textit{legal/noncompliance} risk categories (Section 4). We demonstrate that extractions from single vs. multi-sentence distance groups outperform the baseline by statistically significant margins. We observe that as coverage of the taxonomy grows, the preference for multi-sentence extractions increases, but entity-risk relationship becomes more indirect. We discuss these results relative to the user, information recall minimization and taxonomy management (Section 5). We conclude with further considerations and extensions of the system (Section 6). 

\section{Related Work}

A central focus in risk mining is on the ML classification of risk from non-risk and contributing features - e.g. textual association with stock price movement and financial documents (\citet{KoganEtAl2009}, \citet{LuEtAl2009a}, \citet{GrothMuntermann2011}, \citet{TsaiWang2012}, and \citet{DasguptaEtAl2016}); sentiment of banking risks \cite{NoppHanbury2015}; and risk in news (\citet{LuEtAl2009b} and \citet{NugentEtAl2017}). Heuristic approaches also exist - e.g., risk taxonomy built with seed patterns used in earnings report altering \citet{LeidnerSchilder2010} and company risk rankings based on supply chain analysis \cite{CarstensEtAl2017}. Additional research is focused on curation and management of risk taxonomies (e.g. ontology merging \cite{SubramaniamEtAl2010}, crowdsourcing \cite{MengEtAl2015} and paraphrase detection \cite{PlachourasEtAl2018}).

Our system is closest to the system proposed in \citet{NugentLeidner2017}, which starts with an risk taxonomy (following \citet{LeidnerSchilder2010}) to be searched in news data. All possible extracts are paired with all possible companies and, based on an SVM model built from annotated data, tuples are classified and stored for risk analyst review. Our system similarly focuses on entity-risk relations for analyst review, but deviates in two key ways: (a) because of the specificity of the initial taxonomy, extracts are assumed to express some degree of entity-risk relationship, obviating the need for a risk classifier; and (b) extracts are based on ``shallow" surface parsing rather than deeper morpho-syntactic parsing. These deviations will be revisited in Section 5.

\section{System}

\begin{figure*}[t]
	\includegraphics[width=\linewidth]{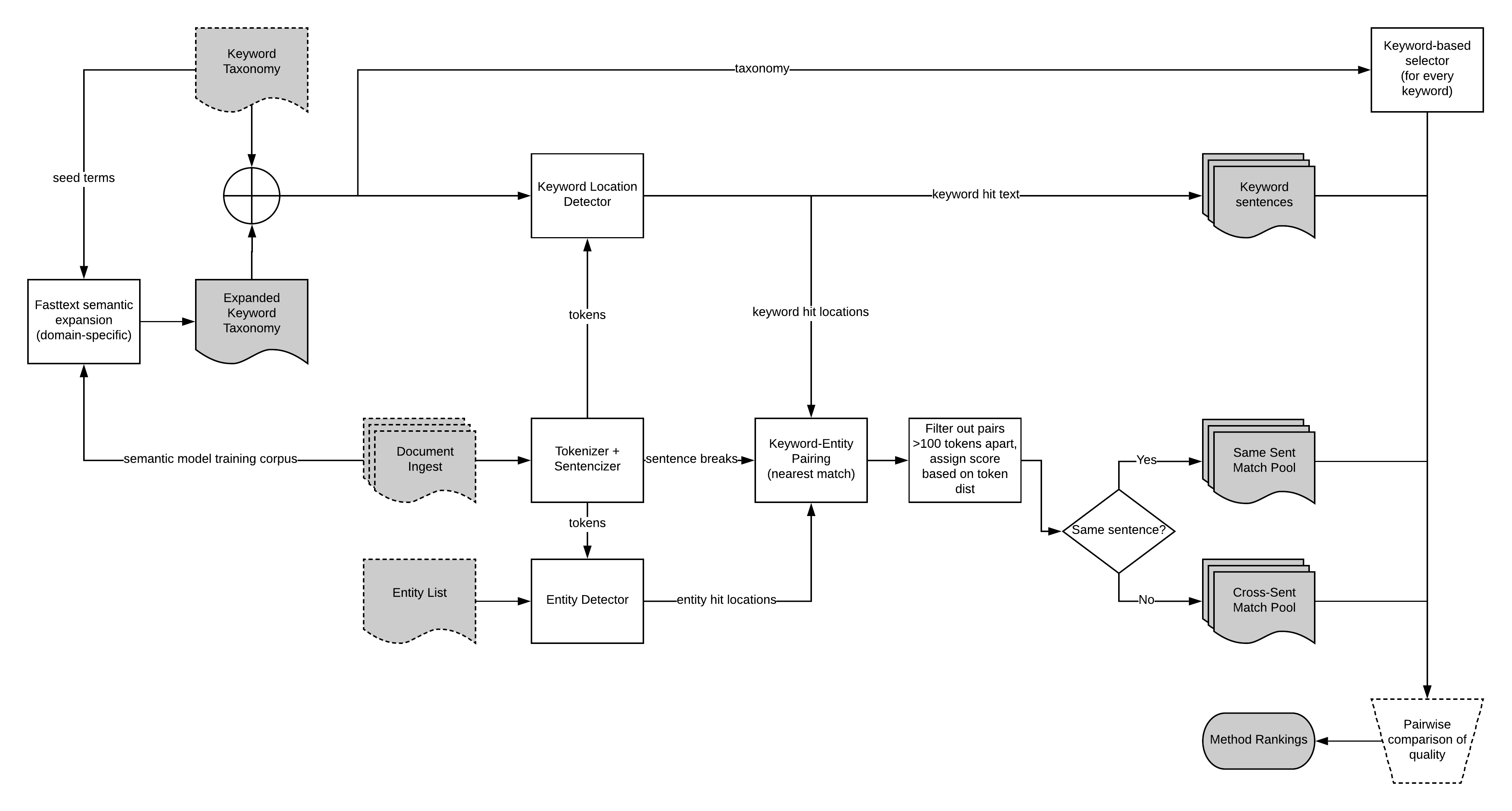}
	\caption{System Diagram. Unshaded elements represent processes, while shaded nodes represent data input or products. Solid-bordered nodes are the system's automatic processors or output, and dashed-bordered elements are data sources provided to the system or expert knowledge. The architecture is designed such that each processor element can operate independently (without shared memory) and is NLP platform-independent.}
\end{figure*}

Our system (Figure 1) is a custom NLP processing pipeline capable of the ingesting and analyzing hundreds of thousands of text documents. The system consists of four components:

\footnotesize
\begin{enumerate}
	\item \textbf{Document Ingest and Processing:} Raw text documents are read from disk then tokenized, lemmatized, and sentencized. 
	\item \textbf{Keyword/Entity Detection:} Instances of both keywords and entities are identified in the processed text, and each risk keyword occurrence is matched to the nearest entity token.
	\item \textbf{Match Filtering and Sentence Retrieval:} Matches within the documents are filtered and categorized by pair distance and/or sentence co-occurrence, and the filtered sentences are retrieved for context.
	\item \textbf{Semantic Encoding and Taxonomy Expansion:} A semantic vectorization algorithm is trained on domain-specific text and used to perform automated expansion of the keyword taxonomy.
\end{enumerate}

\normalsize
\noindent This design architecture allows for significant customization, high throughput, and modularity for uses in experimental evaluation and deployment in production use-cases. We built the system to support decentralized or streaming architectures, with each document being processed independently and learning systems (specifically at the semantic encoding/expansion steps) configured for continuous learning or batch model training. 

\subsection{Document Ingest and Processing}

We leverage \textit{spaCy} (Version 2.0.16, https://spacy.io) \cite{HonnibalJohnson2015} as the document ingest and low-level NLP platform for this system. This choice was influenced by spaCy's high speed parsing \cite{ChoiEtAl2015}, out-of-the-box parallel processing, and Python compatibility, spaCy's default NLP pipeline runs \textit{tokenizer $\rightarrow$ part-of-speech tagger $\rightarrow$ dependency parser $\rightarrow$ named entity recognizer}.

We used the sentence breaks found by the dependency parser to annotate each of the found keyword-entity pairs as being either in the same or different sentences. A dependency-based sentencizer is preferred to a simpler stop-character based approach due to the unpredictable formatting of certain domains of text - e.g. web-mined news and regulatory filings.

spaCy's \textit{pipe()} function, allows for a text generator object to be provided, and takes advantage of multi-core processing to parallelize batching. In this implementation, each processed document piped in by spaCy is converted to its lemmatized form with sentence breaks noted so that sentence and multi-sentence identification of keyword/entity distances can be captured.

\subsection{Keyword/Entity Detection}
In the absence of intervening information or a more sophisticated approach to parsing, the mention of an entity and risk keyword within a phrase or sentence is the most morpho-syntactically, semantically and and pragmatically coherent of the relationship. For example, (3) describes the entity \textbf{Verizon} and its litigation risk associated with lawsuit settlement (keywords being \underline{\smash{settle}} and \underline{\smash{lawsuit}}).\\ 

\noindent \footnotesize (3) \textit{In 2011, \textbf{Verizon} agreed to pay \$20 million to \underline{\smash{settle}} a class-action \underline{\smash{lawsuit}} by the federal Equal Employment Opportunity Commission alleging that the company violated the Americans with Disabilities Act by denying reasonable accommodations for hundreds of employees with disabilities.}\\

\normalsize 
Returning the entire sentence yields additional information - the lawsuit is \textit{class-action} and the complaint allegation is that \textbf{Verizon} ``\textit{denied reasonable accommodations for hundreds of employees with disabilities.}" The system detection process begins by testing for matches of each keyword with each entity, for every possible keyword-entity pairing in the document. \textbf{Algorithm 1} provides the simplified pseudocode for this process.

\begin{algorithm}[h]
	\small
	\caption{Entity-Keyword Pairing}
	\begin{algorithmic}
	\REQUIRE taxonomy and entities lists 
	\FOR{keyword in taxonomy}
	\FOR{entity in entities}
	\STATE keywordLocs = findLocs(keyword)
	\STATE entityLocs = findLocs(entity)
	\FOR{kLoc in keywordLocs}
	\STATE bestHit = findClosestPair(kLoc, entityLocs)
	\STATE results.append((keyword, entity, bestHit))
	\ENDFOR
	\ENDFOR
	\ENDFOR
	\RETURN findClosestPair (two token indicies)
	\end{algorithmic}
\end{algorithm}

For every instance of every keyword, the nearest instance of every available entity is paired - regardless of whether it precedes or proceeds the keyword. Furthermore, an entity may be found to have multiple risk terms associated with it, but each \textit{instance} of a risk term will only apply itself to the closest entity - helping to prevent overreaching conclusions of risk while maintaining system flexibility.\\

\noindent \footnotesize (4) \textit{McDonald says this treatment violated the terms of a \underline{\smash{settlement}} the company reached a few years earlier regarding its treatment of employees with disabilities. In 2011, \textbf{Verizon} agreed to pay \$20 million to settle a class-action lawsuit by the federal Equal Employment Opportunity Commission ....}\\

\normalsize
For example, (4) extends the extract of (3) to the prior contiguous sentence which contains \textit{settlement}. This extension provides greater context for Verizon's lawsuit. (4) is actually background for a larger proposition being made in the document that Verizon is in violation of settlement terms from a \textit{previous} lawsuit.

\normalsize 
The system's token distance approach promotes efficiency and is preferable to more complex NLP - e.g., chunking or co-reference resolution. Nonetheless, this flexibility comes at a computational cost: a total of $ (m \cdot a) \times (n \cdot b) $ comparisons must be made for each document, where \textit{m} is the number of keyword terms across all taxonomic categories, \textit{a} the average number of instances of each keyword per document, \textit{n} the number of entities provided, and \textit{b} the average number of entity instances per document. Changing any single one of these variables will result in computational load changing with $\mathcal{O}(n)$ complexity, but their cumulative effects can quickly add up. For parallelization purposes, each keyword is independent of each other keyword and each entity is independent of each other entity. This means that in an infinitely parallel (theoretical) computational scheme, the system runs on $\mathcal{O}(a \times b)$, which will vary as a function of the risk and text domains.

\begin{table*}
	\small
	\centering
	\begin{tabular}{|l|l|l|}
		\hline
		\textbf{Risk Category}&\textbf{Keyword Seed Taxonomy} & \textbf{Enriched Keyword Taxonomy}\\
		\hline
		\textit{\textbf{Cybersecurity}} & \textit{n=26} & \textit{n=123}\\
		\hline
		&\textit{cybercrime, hack, DDOS, antivirus,} & \textit{4front security, cyber deterence,}\\
		&\textit{data breach, ransomware, penetration, ...} & \textit{cloakware, unauthenticated, ...}\\
		\hline
		\textit{\textbf{Terrorism}} & \textit{n=37}& \textit{n=147}\\
		\hline
		&\textit{terrorism, bio-terrorism, extremist,}& \textit{anti-terrorism, bomb maker, explosives,}\\
		&\textit{car bombing, hijack, guerrilla, ...}& \textit{hezbollah, jihadi, nationalist, ...}\\
		\hline
		\textit{\textbf{Legal}} & \textit{ n=38}& \textit{n=162}\\
		\hline
		&\textit{litigation, indictment, allegation,}&\textit{appropriation, concealment, counter suit,}\\
		&\textit{failure to comply, sanctions violations, ...}&\textit{debtor, expropriation, issuer, ...}\\
		\hline
	\end{tabular}
	\caption{Sample risk terms from the seed and expanded sets.}
\end{table*}

\subsection{Match Filtering and Sentence Retrieval}

The system has by this point completed the heart of document processing and risk identification. The next component seeks to (a) filter away results unlikely to hold analytic value; and (b) identify the hits as being either single sentence or multi-sentence using the sentence information generated by the spaCy dependency parse. The first of these goals is achieved with a simple hit distance cutoff wherein any keyword-entity pair with more than a certain count of intervening tokens is discarded. The setting of a hard cutoff improves keyword-entity spans by not including cross-document matches for large documents. Once filtering is complete, the system uses document sentence breaks identified by spaCy to the sentence membership of each keyword and entity for each pairing and, ultimately, whether they belong to the same sentence. 

\subsection{Encodings}

Our system automates term expansion by using similarity calculations of semantic vectors. We generated these vectors by training a \textit{fastText} (https://fasttext.cc/) skipgram model, which relies on words and subwords from the same data source as the initial extractions \cite{BojanowskiEtAl2017}. This ensures that domain usage of language is well-represented, and any rich domain-specific text may be used to train semantic vectors (\textit{see generally}, \citet{MikolovEtAl2013}). We chose fastText was chosen for this project because of its light weight, open source availability, efficient single-machine performance, and output of vector models for use in Python.

For each taxonomic risk term encountered, fastText searches the model vocabulary for the minimized normalized dot product $\frac{r \cdot w}{\| r \| w \|}$ (a basic similarity score found in the fastText codebase), and returns the top-scoring vocabulary terms as candidates for taxonomic expansion.

\section{Experiment}

To test the performance of the system, we designed an experiment comparing the performances of systems using single-sentence risk detection and systems only using multi-sentence risks. This measures the performance of purely multi-sentence hits against purely single-sentence hits. In addition, a baseline system was tested that detected only risk terms without searching for corresponding entities. Taken together, the three hypotheses tested are as follows:\\

\footnotesize
\hspace{5mm}$ H_{1}: p_{multi} > p_{base} $, $ H_{\varnothing}: p_{multi} = p_{base} $

\hspace{5mm}$ H_{2}: p_{single} > p_{base} $, $ H_{\varnothing}: p_{single} = p_{base} $

\hspace{5mm}$ H_{3}: p_{multi} > p_{single} $, $ H_{\varnothing}: p_{multi} = p_{single} $\\

\normalsize
$ H_{1}$ and $ H_{2}$ test whether each method of detecting risk co-occurrence with an entity performs better than random chance of selecting a risk term in the document and its association with the entity. $ H_{3}$ tests whether the distance-based measure in the system outperforms a sentential approach.

\subsection{Data Processing}

A virtualized Ubuntu 14.04 machine with 8 vCPUs - running on 2.30GHz Intel Xeon E5-2670 processors and 64GB RAM - was chosen to support the first experiment.

The names of the top Fortune 100 companies from 2017 (http://fortune.com/fortune500/2017/) were fed as input into a proprietary news retrieval system for the most recent 1000 articles mentioning each company. Ignoring low coverage and bodiless news articles, 99,424 individual documents were returned. Each article was then fed into the system and risk detections found with a distance cutoff of 100 tokens. For each identified risk, whether single or multi-sentence, the system also selected a baseline sentence at random from the corresponding document for pairwise comparison. 

The spaCy dependency parse was the largest bottleneck - expected total runtime for the near 100,000 documents at approximately 7 calendar-days of computation. In the interest of runtime, only the first 21,000 documents read in order of machine-generated news article ID were analyzed. Once all selected documents were processed, we paired single and multi-sentence spans relating to the same risk category, but potentially different entities and documents, for pairwise evaluations.

\subsection{Term Expansion}
As summarized in Table 1, starting with manually-created seed terms in each category of risk, encodings were learned using fastText from a concatenation of the news article text using the methodology discussed in Section 3.4. Selecting the top ten most similar terms for each in-vocabulary seed term resulted in an expanded taxonomy, with a 326.31\% increase on average across the three categories. This term expansion not only introduced new vocabulary to the taxonomy but also variants and common misspellings of keywords, which are important in catching risk terms``in the wild". Some cleanup of the term expansion was required to filter out punctuation and tokenization variants. 

\subsection{Evaluation}
Analysts were asked to give their preference for ``System A" or ``System B" or ``Neither" when presented with randomized pairs of output. Percentage preferences for the overarching system and each of six pairings was tested for significance with Pearson's $\chi^2$ using raw counts.

\begin{figure*}[t]
	\includegraphics[width=\linewidth]{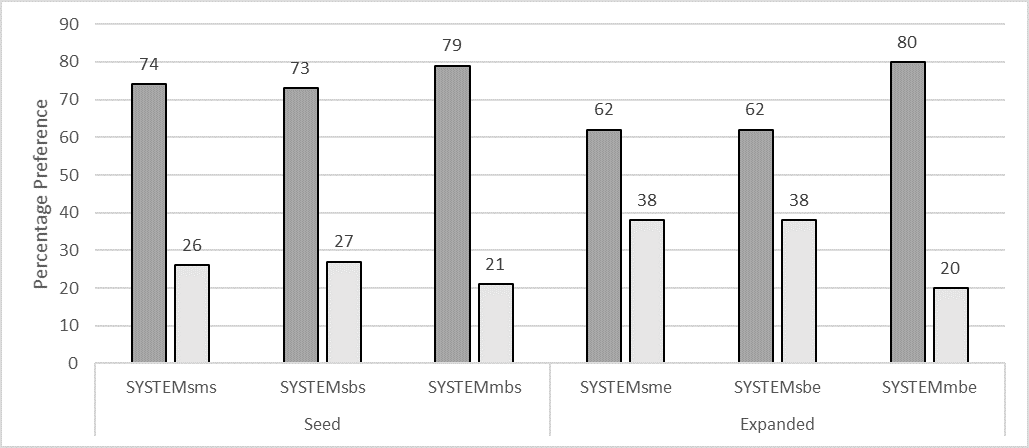}
	\caption{Expert preference ratings. Left and columns are single or multi-sentence as defined in Table 2.}
\end{figure*}

\begin{table}[h]
	\begin{tabular}{|l|c|}
		\hline
		\textbf{System}&\textbf{$\chi^2$}\\
		\hline
		\hline
		Overall & 8.530\\
		single v. multi (seed v. expand) & \textit{p}=0.003\\
		\hline
		\hline
		SYSTEM\textit{sbs} & 28.088\\
		single v. baseline (seed) &  \textit{p}=1.159e-07\\
		\hline
		SYSTEM\textit{mbs} & 25.358\\
		multi v. baseline (seed) &  \textit{p}=4.762e-07\\
		\hline
		SYSTEM\textit{sms} & 37.763\\
		single v. multi (seed) &  \textit{p}=7.99e-10\\
		\hline
		SYSTEM\textit{sbe} & 6.858\\
		single v. baseline (expand) &  \textit{p}=0.008\\
		\hline
		SYSTEM\textit{mbe} & 25.705\\
		multi v. baseline (expand) &  \textit{p}=3.978e-07\\
		\hline
		SYSTEM\textit{sme} & 6.316\\
		single v. multi (expand) &  \textit{p}=0.011\\
		\hline
	\end{tabular}
	\caption{Pearson's $\chi^2$ and \textit{p} values (d.f.=1)}
\end{table}

\section{Results and Discussion}

We collected 4,514 judgments from eight subject matter experts to compute system preferences associated with single, multi- and baseline sentence extractions. Roughly ~28\% of all evaluated extractions (1,266/4,514) received a preference judgment (32\% from the seed set (698/2,198) and 24\% from the expansion set (568/2,316)) where ~72\% received a ``Neither" rating. 

As summarized in Table 2 and Figure 2, all single and multi-sentence extractions across the seed and expansion sets outperform the baseline by statistically significant margins. For the seed set, the single sentence extractions outperform the multi-sentence extractions by a statistically significant margin as well (\textit{p}$\leq$.01). However, for the expansion set, the multi-sentence extractions gain significant ground (26\% to 38\% increase in preference). 

1,283 (28\%) of evaluations were doubly annotated for calculation of Cohen's Kappa \cite{Cohen1960}. Average $\kappa$ for the seed set was 0.284 and 0.144 for the expansion set (which suffers from low sample size). This is uniformly low across all categories, but not an unusual result given the task and the range of analyst expertise. 

\subsection{Discussion}

Our system's distance metric is clearly providing benefit well above the baseline, so we can accept $ H_{1}$ and $ H_{2}$, but given exactness of the seed taxonomy, the high proportion of ``Neither" judgments is surprising. While there are polysemous examples (non-risk category uses of risk keywords), which are minimized in more complex systems, there are observable variations in the directness or indirectness of the entity-risk relationships. For example, \textit{The companies could face a number of lawsuits from Walmart, Target and Kroger}, describes an entity-risk relationship where the entity is suing rather than being sued. The \textit{degree} of the risk varied by analyst and will have to be controlled for in subsequent evaluations and possibly components of the system.

We cannot accept $ H_{3}$ which tests the assumption that multi-sentence returns will have greater analytical utility. However, we observed that as the taxonomy expands, the preference for the multi-sentences increases by 46\% over the single sentence seed set extractions (SYSTEMsms).\footnote{Also note that despite an overall 326\% increase in coverage, the proportion of ``Neither" judgments in a randomly selected set of 100 sentences increases from 52\% to 75\%.} A potential reason for this is that the keywords in the expanded taxonomy exhibit a greater range of specificity as compared to the seed terms. For example, the expanded taxonomy included \textit{require} and \textit{suit} which are more general and given to ambiguity in the \textit{legal/noncompliance} category (the expanded taxonomy also included highly specific keywords - e.g. \textit{foreclosure} and \textit{harassment}). 

To the extent that more general keywords proliferate the taxonomy, multi-sentence information may be more preferred for understanding the entity-risk relationship. This additional information may be preferred in some use cases, but the potential increase via a concomitant encouragement of non-risk relationships would certainly be unwelcome. Accommodating shifts in semantic granularity as the taxonomy expands will likely have to be considered in addition to distance thresholding to tune results.  

\section{Conclusion and Future Work}

We have described a configurable, scalable system for finding entity-event relations across large data sets. Addressing observed drawbacks should be done so relative to maintaining flexibility for analyst users - possibilities include summarization of results for better presentation, alternative source data at the direction of the analyst for given risk categories, token distance thresholding and considerations of semantic granularity. 

While designed for high-coverage use cases in risk mining, our system could be used for any data, entities and taxonomies to support generalized entity-\textit{x} relationship reporting. Further, while the current system seeks to be low complexity (shallow parsing and no risk classifier), the system's modular design can facilitate any number of additions for customized use cases and integrations, and performance comparisons to similar systems. 




\section*{Acknowledgments}
Thank you to anonymous reviewers from NAACL Industry Track as well as Matt Machado, Nathan Maynes, and Scott McFadden for support and feedback. Thank you also to extraction output evaluators Alli Zube, Jason Sciarotta, Jeff Earnest, Jonathan Feng, Larry Fowlkes, Regina Reese, Romel Lira and Sean Ahearn.

\bibliography{naaclhlt2019}

\begin{thebibliography}{19}
\expandafter\ifx\csname natexlab\endcsname\relax\def\natexlab#1{#1}\fi

\bibitem[{Bojanowski et~al.(2017)Bojanowski, Grave, Joulin, and
  Mikolov}]{BojanowskiEtAl2017}
Piotr Bojanowski, Edouard Grave, Armand Joulin, and Tomas Mikolov. 2017.
\newblock \href {http://aclweb.org/anthology/Q17-1010} {Enriching word vectors
  with subword information}.
\newblock \emph{Transactions of the Association for Computational Linguistics},
  5:135--146.

\bibitem[{Carstens et~al.(2017)Carstens, Leidner, Szymanski, and
  Howald}]{CarstensEtAl2017}
Lucas Carstens, Jochen~L. Leidner, Krzysztof Szymanski, and Blake Howald. 2017.
\newblock \href {https://doi.org/10.1007/978-3-319-58451-5\_2} {Modeling
  company risk and importance in supply graphs}.
\newblock In \emph{The Semantic Web - 14th International Conference, {ESWC}
  2017, Portoro{\v{z}}, Slovenia, May 28 - June 1, 2017, Proceedings, Part
  {II}}, pages 18--32.

\bibitem[{Choi et~al.(2015)Choi, Tetreault, and Stent}]{ChoiEtAl2015}
Jinho~D. Choi, Joel Tetreault, and Amanda Stent. 2015.
\newblock \href {http://www.aclweb.org/anthology/P15-1038} {It depends:
  Dependency parser comparison using a web-based evaluation tool}.
\newblock In \emph{Proceedings of the 53rd Annual Meeting of the Association
  for Computational Linguistics}, pages 387--396.

\bibitem[{Cohen(1960)}]{Cohen1960}
Jacob Cohen. 1960.
\newblock \href {https://doi.org/10.1177/001316446002000104} {A coefficient of
  agreement for nominal scales}.
\newblock \emph{Educational and Psychological Measurement}, 20(1):37--46.

\bibitem[{Dasgupta et~al.(2016)Dasgupta, Dey, Dey, and Saha}]{DasguptaEtAl2016}
Tirthankar Dasgupta, Lipika Dey, Prasenjit Dey, and Rupsa Saha. 2016.
\newblock \href {http://aclweb.org/anthology/C16-2038} {A framework for mining
  enterprise risk and risk factors from text}.
\newblock In \emph{Proceedings of COLING 2016, the 26th International
  Conference on Computational Linguistics: System Demonstrations}, pages
  180--184.

\bibitem[{Groth and Muntermann(2011)}]{GrothMuntermann2011}
Sven~S. Groth and Jan Muntermann. 2011.
\newblock \href {https://doi.org/10.1016/j.dss.2010.08.019} {An intraday market
  risk management approach based on textual analysis}.
\newblock \emph{Decision Support Systems}, 50(4):680--691.

\bibitem[{Honnibal and Johnson(2015)}]{HonnibalJohnson2015}
Matthew Honnibal and Mark Johnson. 2015.
\newblock \href {http://aclweb.org/anthology/D15-1162} {An improved
  non-monotonic transition system for dependency parsing}.
\newblock In \emph{Proceedings of the 2015 Conference on Empirical Methods in
  Natural Language Processing}, pages 1373--1378, Lisbon, Portugal. Association
  for Computational Linguistics.

\bibitem[{Kogan et~al.(2009)Kogan, Dimitry~Levin, Sagl, and
  Smith}]{KoganEtAl2009}
Shimon Kogan, Bryan R.~Routledge Dimitry~Levin, Jacob~S. Sagl, and Noah~A.
  Smith. 2009.
\newblock \href {http://www.aclweb.org/anthology/N09-1031} {Predicting risk
  from financial reports with regression}.
\newblock In \emph{Proceedings of the 2009 Annual Conference of the North
  American Chapter of the ACL (NAACL-HLT)}, pages 272--280.

\bibitem[{Leidner and Schilder(2010)}]{LeidnerSchilder2010}
Jochen~L. Leidner and Frank Schilder. 2010.
\newblock \href {http://aclweb.org/anthology/P10-4010} {Hunting for the black
  swan: Risk mining from text}.
\newblock In \emph{Proceedings of the ACL 2010 System Demonstrations}, pages
  54--59.

\bibitem[{Lu et~al.(2009a)Lu, Huang, Li, and Chen}]{LuEtAl2009a}
Hsin-Min Lu, Nina Wan-Hsin Huang, Shu-Hsing Li, and Tsai-Jyh Chen. 2009a.
\newblock Risk statement recognition in news articles.
\newblock In \emph{Proceedings of the 2009 Annual Conference of the
  International Conference on Information Systems}, pages 54--59.

\bibitem[{Lu et~al.(2009b)Lu, Huang, Zhang, and Chen}]{LuEtAl2009b}
Hsin-Min Lu, Nina~WanHsin Huang, Zhu Zhang, and Tsai-Jyh Chen. 2009b.
\newblock \href {https://doi.org/10.1007/978-3-642-01393-5_6} {Identifying
  firm-specific risk statements in news articles}.
\newblock In \emph{Intelligence and Security Informatics}, pages 42--53.
  Springer.

\bibitem[{Meng et~al.(2015)Meng, Tong, Chen, and Cao}]{MengEtAl2015}
Rui Meng, Yongxin Tong, Lei Chen, and Caleb~Chen Cao. 2015.
\newblock \href {https://doi.org/10.1109/ICDM.2015.77} {Crowdtc: Crowdsourced
  taxonomy construction}.
\newblock In \emph{2015 IEEE International Conference on Data Mining}, pages
  913--918.

\bibitem[{Mikolov et~al.(2013)Mikolov, Chen, Corrado, and
  Dean}]{MikolovEtAl2013}
Tomas Mikolov, Kai Chen, Greg Corrado, and Jeffrey Dean. 2013.
\newblock \href {https://doi.org/10.1162/neco\_a\_01017} {Efficient estimation
  of word representations in vector space}.
\newblock \emph{arXiv preprint arXiv:1301.3781}.

\bibitem[{Nopp and Hanbury(2015)}]{NoppHanbury2015}
Clemens Nopp and Allan Hanbury. 2015.
\newblock \href {https://doi.org/10.18653/v1/D15-1071} {Detecting risks in the
  banking system by sentiment analysis}.
\newblock In \emph{Proceedings of the 2015 Conference on Empirical Methods in
  Natural Langauge Processing}, pages 591--600.

\bibitem[{Nugent and Leidner(2016)}]{NugentLeidner2017}
Timothy Nugent and Jochen~L. Leidner. 2016.
\newblock \href {https://doi.org/10.1109/ICDMW.2016.0191} {Risk mining:
  Company-risk identification from unstructured sources}.
\newblock In \emph{{IEEE} International Conference on Data Mining Workshops,
  {ICDM} Workshops 2016, December 12-15, 2016, Barcelona, Spain.}, pages
  1308--1311.

\bibitem[{Nugent et~al.(2017)Nugent, Petroni, Raman, Carstens, and
  Leidner}]{NugentEtAl2017}
Timothy Nugent, Fabio Petroni, Natraj Raman, Lucas Carstens, and Jochen~L.
  Leidner. 2017.
\newblock \href {https://doi.org/10.1109/BigData.2017.8258374} {A comparison of
  classification models for natural disaster and critical event detection from
  news}.
\newblock In \emph{2017 IEEE International Conference on Big Data (Big Data)},
  pages 3750--3759.

\bibitem[{Plachouras et~al.(2018)Plachouras, Petroni, Nugent, and
  Leidner}]{PlachourasEtAl2018}
Vassilis Plachouras, Fabio Petroni, Timothy Nugent, and Jochen~L. Leidner.
  2018.
\newblock \href {http://www.aclweb.org/anthology/N18-2051} {A comparison of two
  paraphrase models for taxonomy augmentation}.
\newblock In \emph{Proceedings of the 2018 Annual Conference of the North
  American Chapter of the ACL (NAACL-HLT)}, pages 315--320.

\bibitem[{Subramaniam et~al.(2010)Subramaniam, Nanavati, and
  Mukherjea}]{SubramaniamEtAl2010}
L.~Venkata Subramaniam, Amit~Anil Nanavati, and Sougata Mukherjea. 2010.
\newblock \href {https://doi.org/10.1109/TKDE.2009.189} {Enriching one taxonomy
  using another}.
\newblock In \emph{2010 IEEE Transactions on Knowledge and Data Engineering},
  pages 913--918.

\bibitem[{Tsai and Wang(2012)}]{TsaiWang2012}
Ming-Feng Tsai and Chuan-Ju Wang. 2012.
\newblock \href {http://www.aclweb.org/anthology/C12-3056} {Visualization on
  financial terms via risk ranking from financial reports}.
\newblock In \emph{Proceedings of COLING 2012}, pages 447--452.

\end{thebibliography}
\bibliographystyle{acl_natbib}

\end{document}